%% file: main.tex
\documentclass[10pt,letterpaper]{article}

\usepackage{cogsci}

\usepackage{pslatex}
\usepackage{apacite}
\usepackage{float} %

\usepackage{graphicx} %

\usepackage{xcolor} %
\usepackage{ascmac} %
\usepackage{url}
\usepackage{array}     %
\usepackage{booktabs}  %
\usepackage{adjustbox} %
\usepackage{caption}
\usepackage{enumitem}  %

\newif\ifcogscifinal %
\cogscifinaltrue

\title{Do Large Vision-Language Models Distinguish \\between the Actual and Apparent Features of Illusions?}

\author{%
  Taiga Shinozaki${}^{2,1}$ \ %
  Tomoki Doi${}^{1}$ \ %
  Amane Watahiki${}^{1}$ \ %
  Satoshi Nishida${}^{3}$ \ %
  Hitomi Yanaka${}^{1,4}$ \\[1ex]
  ${}^{1}$The University of Tokyo %
  ${}^{2}$Keio University %
  ${}^{3}$CiNet, NICT %
  ${}^{4}$Riken \\[1ex]
  \texttt{snzktig@keio.jp, amanew@g.ecc.u-tokyo.ac.jp, s-nishida@nict.go.jp} \\
  \texttt{\{doi-tomoki701, hyanaka\}@is.s.u-tokyo.ac.jp}
}

\begin{document}

\maketitle

\begin{abstract}

Humans are susceptible to optical illusions, which serve as valuable tools for investigating sensory and cognitive processes. 
Inspired by human vision studies, research has begun exploring whether machines, such as large vision language models (LVLMs), exhibit similar susceptibilities to visual illusions. 
However, studies often have used non-abstract images and have not distinguished actual and apparent features, leading to ambiguous assessments of machine cognition.  
To address these limitations, we introduce a visual question answering (VQA) dataset, categorized into genuine and fake illusions, along with corresponding control images.
Genuine illusions present discrepancies between actual and apparent features, whereas fake illusions have the same actual and apparent features even though they look illusory due to the similar geometric configuration. 
We evaluate the performance of LVLMs for genuine and fake illusion VQA tasks and investigate whether the models discern actual and apparent features. 
Our findings indicate that although LVLMs may appear to recognize illusions by correctly answering questions about both feature types, they predict the same answers for both Genuine Illusion and Fake Illusion VQA questions. This suggests that their responses might be based on prior knowledge of illusions rather than genuine visual understanding. 
The dataset is available at \url{https://github.com/ynklab/FILM}

\end{abstract}

\section{Introduction}
\label{sec:intr}

Humans are susceptible to optical illusions.
For example, in image A in Figure \ref{fig:illusions_ab}, the two red circles are the same size, but the left one appears bigger to humans.
Illusions are used to investigate human vision because they allow us to isolate sensory and cognitive processes~\cite{day1984nature}.

Motivated by studies of human vision, researchers have recently begun investigating machine vision using illusions to determine whether machines, such as deep neural networks (DNNs) and large vision language models (LVLMs), are susceptible to illusions, and if so, whether their susceptibility is similar to that of humans.
This line of inquiry is not merely of theoretical interest.
Because they are language-based models, LVLMs are prone to over-reliance on linguistic priors, potentially ignoring visual input altogether~\cite{Guan_2024_CVPR}.
Illusions provide useful stress tests for determining whether a model truly integrates visual context or simply mimics linguistic patterns.
Practically, understanding illusion could improve human-AI interaction~\cite{zhang2023grounding}.
If machines are susceptible to illusions in a similar way to humans, machines can understand human instructions about illusion images better.
For example, in image A in Figure \ref{fig:illusions_ab}, machines that have human-like susceptibility can understand which circle a human refers to when they say ``the larger red circle.''
In addition, even if a machine is not susceptible to illusions, it can understand what humans mean if the machine understands how illusions appear to humans.
Therefore, it is also important to determine whether a machine that appears to be susceptible to illusions is really susceptible to them, and whether it understands how illusions appear to humans.
\input{figure/illusionsAB}

DNNs are susceptible to visual illusions, similar to humans~\cite{afifi2019else, benjamin2019shared, gomez2019convolutional, sun2021imagenet}, and visual illusions can mislead LVLMs just as they do human observers~\cite{Guan_2024_CVPR, shahgir2024illusionvqa, zhang2023grounding}. However, models sometimes show responses that diverge from human perception~\cite{ullman2024illill}.
These findings provide insight into the capabilities and limitations of machine perception.

\input{table/illque}

However, the experimental settings of previous studies have two problems in evaluating the illusion recognition abilities of LVLMs accurately, leaving ambiguity in the experimental results.
First, some studies use non-abstract images that create ambiguity in the models' responses \cite{Guan_2024_CVPR, shahgir2024illusionvqa}.
Although abstract images represent two-dimensional abstract shapes, non-abstract images represent objects in three-dimensional space
(Figure \ref{fig:illusions_ab}).
Questions about non-abstract images may be interpreted in two ways.
For example, consider the question ``Which car is the largest?'' about image B in Figure \ref{fig:illusions_ab}.
If this question is interpreted as referring to the image itself, the correct answer would be ``All the cars are the same size.''
A previous study defined this as the correct answer~\cite{shahgir2024illusionvqa}.
However, if the question is interpreted as referring to the objects represented by the image, that is, the cars in the real-world scenario, the correct answer would be ``the car farthest from the observer is the largest.''
This is because, in the real-world scenario, if an object farther from the observer occupies the same size area in the visual field as an object closer to the observer, the farther object must be physically larger.
Therefore, either response can be interpreted as correct, and regardless of which answer a model provides, ambiguity remains in evaluating the model's ability.

The second is that there is further ambiguity in previous studies because they do not distinguish between the \textit{actual} and \textit{apparent features} of the images~\cite{Guan_2024_CVPR, shahgir2024illusionvqa, zhang2023grounding}.
Actual features refer to the features that the image actually has, whereas apparent features refer to the features the image appears to have.
In normal images, apparent features are the same as the actual features.
However, in illusion images, the two are different.
For example, in image A in Figure \ref{fig:illusions_ab}, the left red circle is the same size as the right red circle, although the left circle appears larger than the right circle.
Thus, for the left circle ``being the same size as the right red circle'' is the actual feature, and ``being larger than the right red circle'' is the apparent feature.
Without distinguishing feature types when asking questions, it remains unclear which feature an LVLM is focusing on, regardless of its responses. 
This ambiguity complicates the evaluation of LVLMs' understanding of illusions. 

This study evaluates LVLMs' recognition of illusions more accurately by removing these ambiguities through constructing a dataset of abstract images and prompting the model to distinguish between the actual and apparent features with two types of illusions, genuine and fake illusions.
Genuine illusions have apparent and actual features that are different, whereas fake illusions look like genuine illusions, but their apparent and actual features are the same (Table \ref{tab:illque}).
Using these illusions, we evaluate four LVLMs' recognition of the difference between apparent and actual features.
The results suggest that although LVLMs appear to understand illusions like humans (i.e., to recognize their actual and apparent features) at first glance, their outputs may be based on prior knowledge about illusions and the models may not visually recognize the illusion images correctly.

\section{Related Work}
DNNs and LVLMs are susceptible to visual illusions. DNNs can replicate classical illusions in ways that parallel human errors~\cite{afifi2019else, benjamin2019shared, gomez2019convolutional, sun2021imagenet}, and LVLMs also fall for such illusions~\cite{Guan_2024_CVPR, shahgir2024illusionvqa}. Larger models may even exhibit stronger effects, suggesting closer perceptual alignment with humans~\cite{zhang2023grounding}.

Among recent efforts to evaluate illusion perception in LVLMs, Ullman's work focuses on a specific class of fake illusions, termed ``illusion-illusions''~\cite{ullman2024illill}.
Although the work does not address the ambiguities discussed above, it presents the first systematic evaluation of fake illusions in LVLMs, suggesting that even state-of-the-art models may respond to these illusions in ways that differ from humans.
However, as with other studies, the evaluation remains inconclusive because it involves only machine outputs and no data from human participants.
Moreover, the model responses to open-ended questions are interpreted solely by the author.
By contrast, our approach introduces two key improvements.
First, we conducted parallel experiments with LVLMs and human participants, using identical visual questions to enable direct comparison.
Second, we used a forced-choice visual question answering (VQA) format with fixed answer options (Table\ref{tab:illque}), allowing for fully automatic and reproducible scoring. 

\section{Method}
\label{sec:meth}

\subsection{Task}
To evaluate the illusion recognition ability of LVLMs precisely, we created three VQA tasks, each of which served a distinct purpose: 
i) the Genuine Illusion VQA task, which required answering questions about the features of genuine illusion images;
ii) the Fake Illusion VQA task, which required answering questions about the features of fake illusion images; and
iii) the Control VQA task, which required answering questions about the features of images from which the illusion-inducing factors were removed.

\paragraph{Genuine Illusion VQA} 
In the Genuine Illusion VQA task, we presented genuine illusion images that contain ``illusion inducers,'' which are elements within images that cause the apparent features to differ from the actual features. 
For example, in the Ebbinghaus illusion image shown in Table \ref{tab:illque}, the black circles surrounding these red circles serve as an illusion inducer, making the red circle on the left appear larger than that on the right.
We asked questions about the actual features and apparent features (Table \ref{tab:illque}).
This task setting allowed us to evaluate whether an LVLM responds to illusion images in the same way as humans.

\input{figure/control_images}
\input{figure/prompt}
\paragraph{Fake Illusion VQA} 
Even if an LVLM predicts correct answers on the Genuine Illusion VQA task, this does not necessarily indicate that it recognizes the apparent features of the images accurately. 
Instead, the model could rely on its prior knowledge of the specific illusion~\cite{Guan_2024_CVPR}. 
For example, if an LVLM knows that a circle surrounded by smaller circles in the Ebbinghaus illusion appears larger, it can answer that the circle appears larger without genuinely recognizing the illusion.
The Fake Illusion VQA task determined whether an LVLM genuinely recognized illusions.
In this task, we presented fake illusion images, which still contained illusion inducers but were modified so that their actual features matched their apparent features.  
For example, in the fake illusion image shown in Table \ref{tab:illque}, the left red circle appears larger than the right red circle and is actually larger than the right circle. 
We then ask the same questions as in the Genuine Illusion VQA task.
In the Fake Illusion VQA task, only the correct answers to questions about the actual features differed from those in the Genuine Illusion VQA task (indicated by the red text in Table \ref{tab:illque}). 
If an LVLM genuinely recognizes illusions, it should be able to answer questions correctly about the actual features in the Fake Illusion VQA task.

\paragraph{Control VQA} 
In the Control VQA task, we presented images from which illusion inducers were removed (Figure \ref{fig:control}).
Each image corresponded to a genuine or fake illusion image used in the Genuine or the Fake Illusion VQA task, and we asked the same questions.
We refer to these images as control images corresponding to genuine or fake illusion images. 
This task setting allowed us to distinguish the effects of illusion inducers on LVLM responses from those of other factors.
If an LVLM answers correctly in the Control VQA task but incorrectly in the corresponding Genuine or Fake illusion task, illusion inducers are responsible for its mistake.
On the contrary, if an LVLM answers in the same manner in the Control VQA task and another one, the illusion inducers are irrelevant.

\vskip\baselineskip
Each task addressed the apparent features, which are influenced by illusion inducers, and the corresponding actual features. To distinguish the actual and the apparent features of the images clearly, we created two sets of questions for each feature type and explicitly indicated which feature was addressed in each set (Figure \ref{fig:prompt}). 
Although we do not provide independent evidence that LVLMs robustly master the semantic distinction between ``appears'' and ``is,''
this approach allowed us to reduce the ambiguities present in previous studies.

\subsection{Dataset Construction}
\label{subsec:data}

The dataset construction consisted of three steps: i) illusion selection, ii) image creation, and iii) human evaluation.

\paragraph{Illusion Selection} 
In line with the objectives of this study, we selected well-known illusions that satisfy the following criteria:
i) The illusion involved a discrepancy between perceived and actual features, ensuring consistency in the type of illusions considered\footnote{Previous studies~\cite{shahgir2024illusionvqa,ullman2024illill} have also categorized other image types, such as images of impossible objects, as ``illusions.''} and
ii) The illusion can be depicted as an abstract image to avoid ambiguity associated with non-abstract images.
15 types of illusions were selected in this step (Figure \ref{fig:ex_ills}).
\input{figure/illusion_examples}

\paragraph{Image Creation}
For each of the 15 illusion types, we created two genuine illusion images that differed slightly in angle or color but with identical illusion inducers, resulting in a total of 30 images.
This approach allowed us to assess the consistency of LVLMs' outputs when factors unrelated to the illusions are varied.
Additionally, we created 30 fake illusion images, corresponding to each genuine illusion image, and 60 control images corresponding to each of the genuine and fake illusion images. 
In summary, our preliminary dataset comprised 120 images across 15 illusion types: 30 genuine illusion images, 30 fake illusion images, and 60 corresponding control images.
Genuine illusion images were intended to have apparent features that differ from the actual features as perceived by most human observers. 
Fake illusion images were intended to have no difference between apparent and actual features, enabling most observers to infer the actual features correctly. 

\paragraph{Human Evaluation}
To verify that the created images had the intended features, we conducted an online psychological experiment with 52 human participants. 
The experimental protocol was approved by the ethics committees of the authors' institution. 
Informed written consent was obtained from all participants. 
We created a subset of our preliminary dataset for human evaluation by selecting 68 representative images: 17 genuine illusion images, 17 fake illusion images, and 34 control images. 
This subset included all 15 types of illusions from the original dataset.
To ensure that participants distinguished between apparent and actual features, we presented them with the same prompts used for the models (see Task section).
In the evaluation, the most frequent answer for each question was taken as the representative human answer.\footnote{We did not assess participants' prior familiarity with each illusion. Although this may have influenced responses to some extent, our aim was to compare typical human judgments under naturalistic conditions rather than isolate naive perception. 
}

The representative responses confirmed that all images, except one depicting the Ponzo illusion, exhibited the intended features. 
Because the Ponzo illusion image we created failed to deceive the majority of participants,\footnote{An effective Ponzo illusion should cause two lines within the image to appear unequal in length. However, only 28.9\% of participants perceived an apparent difference in the lengths of the lines in our created image.} we determined that this image did not qualify as a valid genuine illusion image.
Thus, we excluded from the dataset two genuine illusion images depicting the Ponzo Illusion,\footnote{This is because two images are created per type of illusion.} as well as their corresponding fake illusion and control images.
For the fake and control images, human responses matched the intended features with 100.0\% consistency.
The final dataset comprised 112 images across 14 illusion types: 28 genuine illusion images, 28 fake illusion images, and 56 corresponding control images.

\input{table/result-g}
\input{table/result-f}
\section{Experiment 1: Genuine Illusion VQA}
\label{sec:ex1}
\subsection{Experimental Setup}
We evaluated four representative LVLMs on the Genuine Illusion VQA task:
GPT-4o,\footnote{\url{https://openai.com/index/GPT-4o-system-card/}}
Claude 3.5,\footnote{\url{https://assets.anthropic.com/m/61e7d27f8c8f5919/original/Claude-3-Model-Card.pdf}} and 
LLaVA-NeXT (72b, 110b) \cite{li2024llavanext-72-110}.
For each model, we used three prompt settings: 
zero-shot prompting, 
one-shot prompting, 
and 
metacognitive prompting.
In one-shot prompting, the model was provided with a single example consisting of an image and its corresponding question-answer pair before making predictions on new inputs. 
The example image was randomly selected from the dataset, ensuring that the example image was different from the target image being evaluated.
In metacognitive prompting, the model was asked to make a systematic series of structured, self-aware evaluations~\cite{wang2024metacog}.
The temperature value was fixed at 0 to ensure eliminating sampling randomness and maximizing reproducibility.

\subsection{Result and Discussion}

We categorized each of the model's responses on the Genuine Illusion VQA task as follows: i) correct answers to both actual and apparent feature questions (``Both Correct''), ii) correct answers to apparent feature questions only (``Only Apparent''), iii) correct answers to actual feature questions only (``Only Actual''), and iv) incorrect answers to both (``Neither Correct'').
We then calculated the proportions of each response type (Table \ref{tab:result-g}).

As a result, LLaVA-NeXT tended to answer incorrectly about either the apparent features or the actual features in most questions in each prompt setting, regardless of the model size.
In contrast, GPT-4o and Claude 3.5 showed high accuracy for both apparent and actual features in each prompt setting.
Both models rarely gave incorrect outputs about actual features and achieved higher Both Correct rates than that of humans (i.e., 62.5\%) in some prompt settings, such as zero-shot and one-shot prompting for GPT-4o and one-shot and metacognitive prompting for Claude 3.5. 
This difference was particularly evident in color-based illusions, where human participants often misidentified the actual features, whereas both models tended to respond correctly to both the apparent and actual features.
Superficially, this appeared to suggest that the models understood the illusions more accurately than humans.
However, this interpretation is contradicted by Experiment 2.

GPT-4o performed best with a zero-shot strategy, whereas Claude 3.5 benefited from one-shot and metacognitive prompting, with metacognitive prompting giving the highest accuracy. 
These results indicated that GPT-4o may be more robust in handling the Genuine Illusion VQA task without additional context, whereas Claude 3.5 improved when given more guidance. 
Although it was hard to see why the performance of GPT-4o decreased, especially in metacognitive prompting, it may be because this strategy distracted the model from focusing on perceptual information from the image by promoting the use of metacognitive reasoning.
These findings require further research to understand better how different prompt strategies affect model performance.

\section{Experiment 2: Fake Illusion VQA}
\label{sec:ex2}

Because GPT-4o and Claude 3.5 showed sufficient performance on the Genuine Illusion VQA task, we evaluated them on the Fake Illusion VQA task using the same settings as the Genuine Illusion VQA task.

We categorized each model's responses on the Fake Illusion VQA task in the same way as for the Genuine Illusion VQA task.
To calculate the proportions of each response type, we considered only the responses to fake illusion images corresponding to genuine illusion images, where each model correctly answered both the actual and apparent features.
However, for human participants, we included all responses.
The results are presented in Table \ref{tab:result-f}.

Unlike the Genuine Illusion VQA task results in Table \ref{tab:result-g}, GPT-4o and Claude 3.5 tended to answer correctly only for the apparent features with each prompt, even though their outputs on the corresponding Genuine Illusion VQA task were all Both Correct.
This means that many responses followed the pattern shown in Figure \ref{fig:model-ans-ex}.
The prevalence of this pattern 
suggested that the LVLMs did not visually understand the apparent features of the images correctly but were merely inferring these features based on prior knowledge about illusions.
This observation was consistent with those of Ullman \citeyear{ullman2024illill}.
That is, models predict answers based on the pre-existing information and patterns about various illusions that the models have learned from their training data rather than the genuine perceptual understanding of the illusions presented in the images.
For example, regarding the question in Figure \ref{fig:model-ans-ex}, the model mistakenly identified the fake illusion image as a genuine illusion image (Ebbinghaus illusion image) and, based on the prior knowledge that ``in the Ebbinghaus illusion, the actual sizes of the two circles are the same,'' concluded that the actual sizes of the two red circles in the fake illusion image are the same.
\input{figure/model-ans-ex}

\section{Experiment 3: Control VQA}
\label{sec:ex3}
We evaluated GPT-4o and Claude 3.5 on the Control VQA task using the same setting as the Genuine Illusion VQA task.
We categorized each model's responses on the Control VQA task in the same way as for the other tasks.
Tables \ref{com_con1} and \ref{com_con2} present the results of the Control VQA tasks, compared with those of the Genuine Illusion VQA and Fake Illusion VQA tasks, respectively, using zero-shot prompting.
\input{table/comparison_control-g}
\input{table/comparison_control-f}

According to the results of the Control VQA task corresponding to genuine illusions, the Both Correct rates of the Genuine Illusion and the Control VQA tasks indicated little difference in the models (Table \ref{com_con1}). 
This suggested that LVLMs are rarely affected by illusion inducers when recognizing genuine illusion images. 
In contrast, the Both Correct rates between the Fake Illusion and the Control VQA tasks indicate a bigger difference in GPT-4o (see Table \ref{com_con2}).
This result suggested that GPT-4o on the Control VQA task corresponding to fake illusions was strongly influenced by illusion inducers. 
This result increased the plausibility of the explanation that GPT-4o's Only Apparent mistakes on the Fake Illusion VQA task were based on prior knowledge of the illusion.
However, because Claude 3.5 made Only Apparent mistakes on the Control VQA task to the same extent as on the Fake Illusion VQA task, the effect of illusion inducers in the fake illusion images on Claude's responses appeared to be limited (Table \ref{com_con2}).
This observation suggested that factors other than illusion inducers, such as the lack of robust recognition for infrequent abstract images, may contribute to Claude's failure on the Fake Illusion VQA task.
It is difficult to interpret why these models make more errors on the Control VQA task corresponding to fake illusions compared with genuine illusions. 
Therefore, some control images may inadvertently function like fake illusions for the models.
One possible explanation is that, in the case of control images of fake illusions, the correct answer is often that the two figures are actually different, whereas the models tend to answer that they are the same, possibly because of a bias toward visual similarity.
However, this phenomenon requires further investigation.

\section{Conclusion}
\label{sec:conc}

We proposed and implemented VQA tasks to assess the ability of LVLMs to recognize illusions, by asking about the apparent and actual features of genuine and fake abstract illusion images. 
This design was intended to address limitations in previous studies and enable a more accurate evaluation of LVLMs' illusion recognition.
The results for the Genuine Illusion VQA task indicated superficially that the LVLMs seemed to understand illusions.
However, the results for the Fake Illusion VQA task suggest that at least GPT-4o may actually respond based on prior knowledge about illusions rather than real-time visual perception, and that illusion inducers are not the sole cause of failure.
As a future direction, examining the internal mechanisms of LVLMs may help determine whether responses are driven by prior knowledge encoded during training rather than a real-time perceptual processing.
Understanding when and why LVLMs align with human perception remains important for improving human-AI interaction.
We believe that our dataset will 
contribute to further research on the perceptual capabilities of LVLMs.

\newpage
\section{Acknowledgments}
This work was supported by JSPS KAKENHI Grant Numbers JP24H00809, 24K22328, and 24KK0189.

\bibliographystyle{apacite}

\setlength{\bibleftmargin}{.125in}
\setlength{\bibindent}{-\bibleftmargin}

\bibliography{custom}

\end{document}

%% file: figure/illusionsAB.tex
\begin{figure}[t!]
    \centering
    \includegraphics[width=8.5cm]{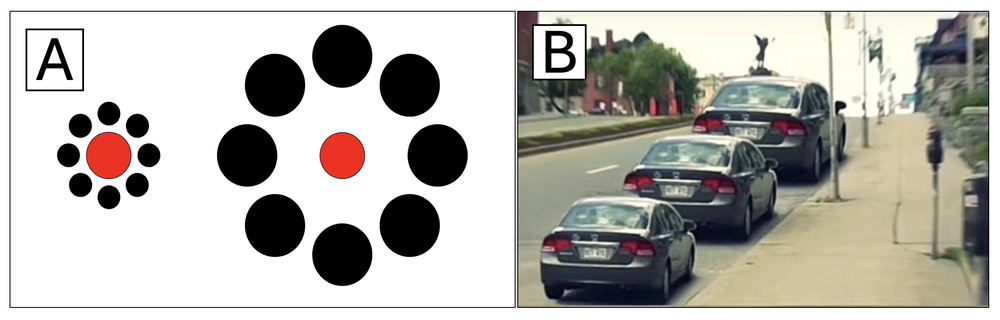}
    \caption{
    (A) Abstract illusion image. 
    Although the two red circles are the same size, the left one appears bigger.
    (B) Non-abstract illusion image.
    Although each car occupies the same size area in the image, the furthest one appears to be the largest~\cite{shahgir2024illusionvqa}.
    }
    \label{fig:illusions_ab}
\end{figure}

%% file: table/illque.tex
\begin{table*}[t]
    \centering
    \caption{Examples of illusion images and questions. These genuine and fake illusion images are for the Ebbinghaus illusion.}
    \label{tab:illque}
    \vskip 0.12in
    \begin{tabular}{|l|l|l|}
        \hline
        Task& 
        \multicolumn{1}{l|}{Genuine Illusion VQA} & Fake Illusion VQA \\ \hline
        Image& 
        \multicolumn{1}{l|}{
        \includegraphics[width=0.3\textwidth]{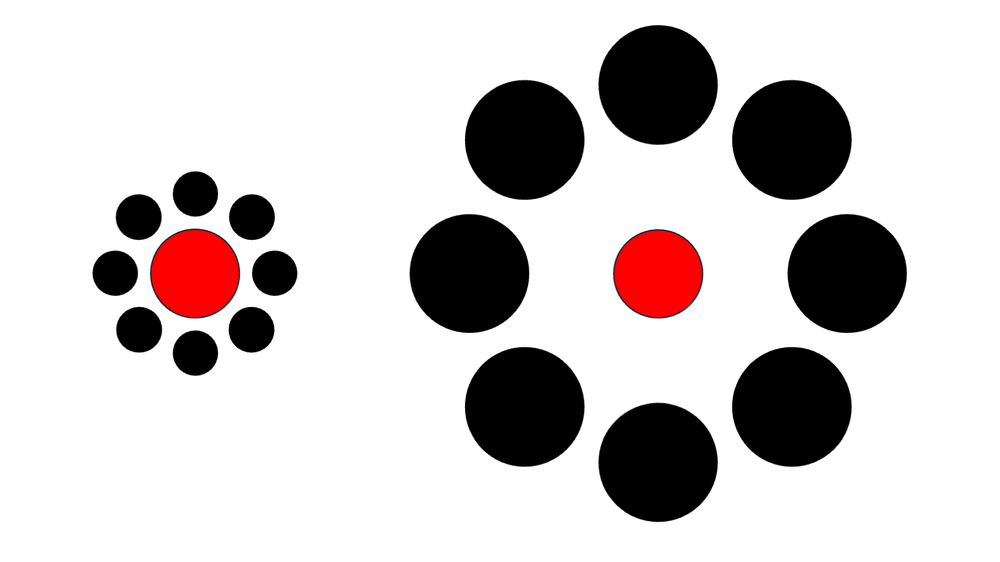}}& \includegraphics[width=0.3\textwidth]{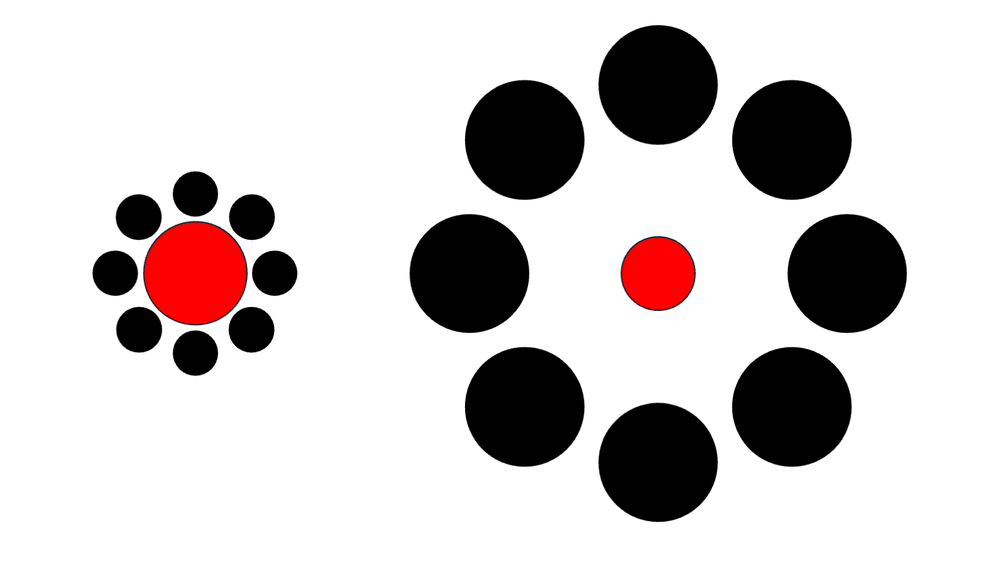} \\ \hline
        Question of actual feature&
        \multicolumn{2}{l|}{\textsf{Which red circle is bigger?}} \\ \hline
        Options of actual feature&
        \multicolumn{2}{l|}{\begin{tabular}[c]{@{}l@{}}
        \textsf{``The left red circle is bigger.'', }\\
        \textsf{``The right red circle is bigger.'', }\\ 
        \textsf{``Both red circles are the same size.''}\end{tabular}} \\ \hline
        Correct answer of actual feature&
        \multicolumn{1}{l|}{\textcolor{red}{\textsf{Both red circles are the same size.}}}&
        \textcolor{red}{\textsf{The left red circle is bigger.}} \\ \hline
        Question of apparent feature&
        \multicolumn{2}{l|}{\textsf{Which red circle appears bigger?}} \\ \hline
        Options of apparent feature&
        \multicolumn{2}{l|}{\begin{tabular}[c]{@{}l@{}}
        \textsf{``The left red circle appears bigger.'', }\\ 
        \textsf{``The right red circle appears bigger.'', }\\ 
        \textsf{``Both red circles appear the same size.''}\end{tabular}} \\ \hline
        Correct answer of apparent feature&
        \multicolumn{2}{l|}{\textsf{The left red circle appears bigger.}} \\ \hline
    \end{tabular}
\end{table*}

%% file: figure/control_images.tex
\begin{figure}[t]
    \centering
    \includegraphics[width=8.5cm]{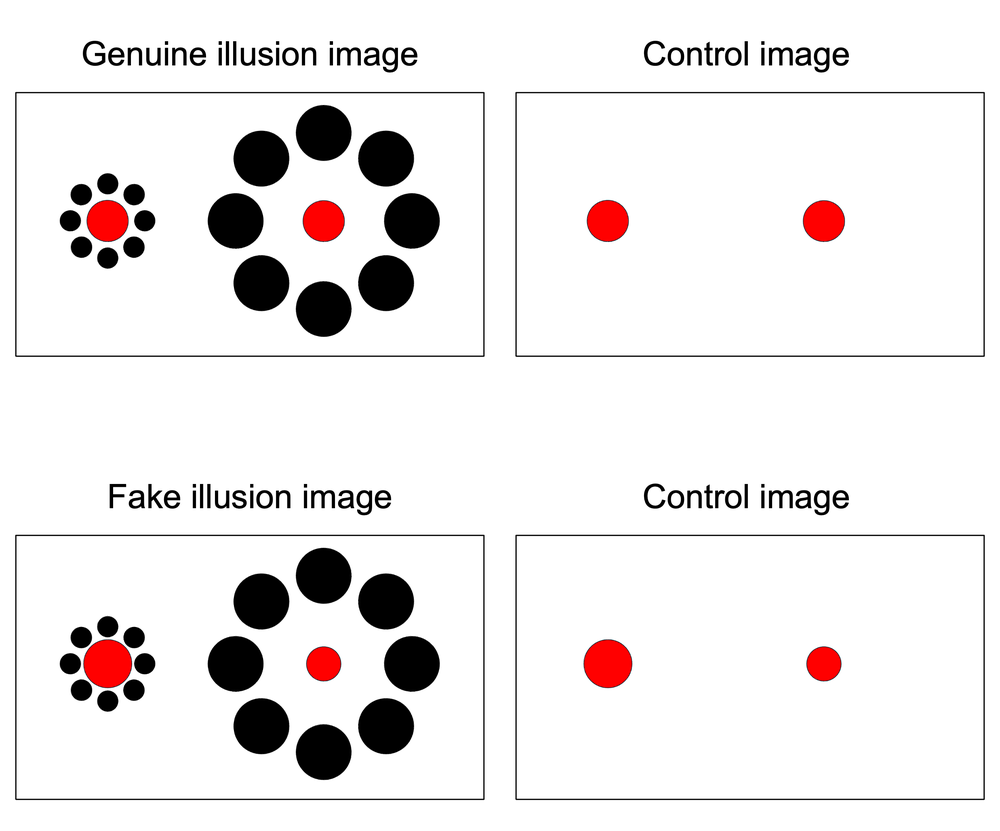}
    \caption{Examples of control images corresponding to each illusion image. Illusion inducers (black circles) are removed from the control images.}
    \label{fig:control}
\end{figure}

%% file: figure/prompt.tex
\begin{figure}[t]
    \centering
    \includegraphics[width=8.5cm]{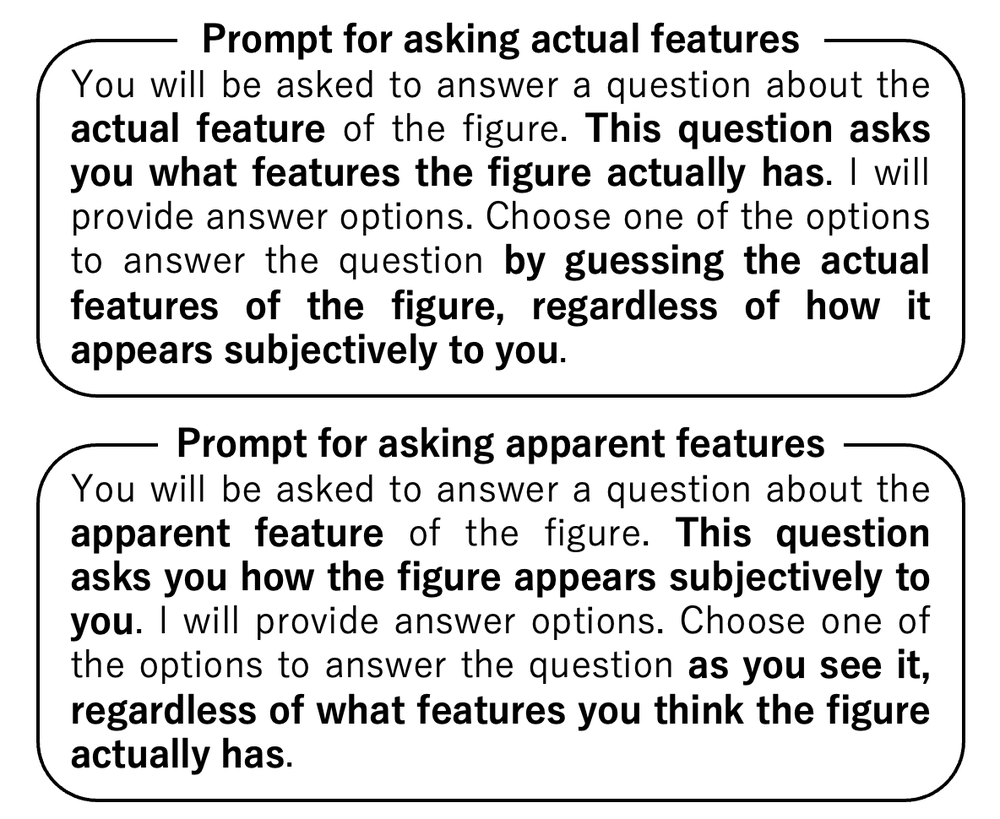}
    \vspace{-1em}
    \caption{
    Prompts used for model evaluation. Bold text indicates the feature types.
    }
    \label{fig:prompt}
\end{figure}

%% file: figure/illusion_examples.tex
\begin{figure}[t]
    \centering
    \includegraphics[width=8.5cm]{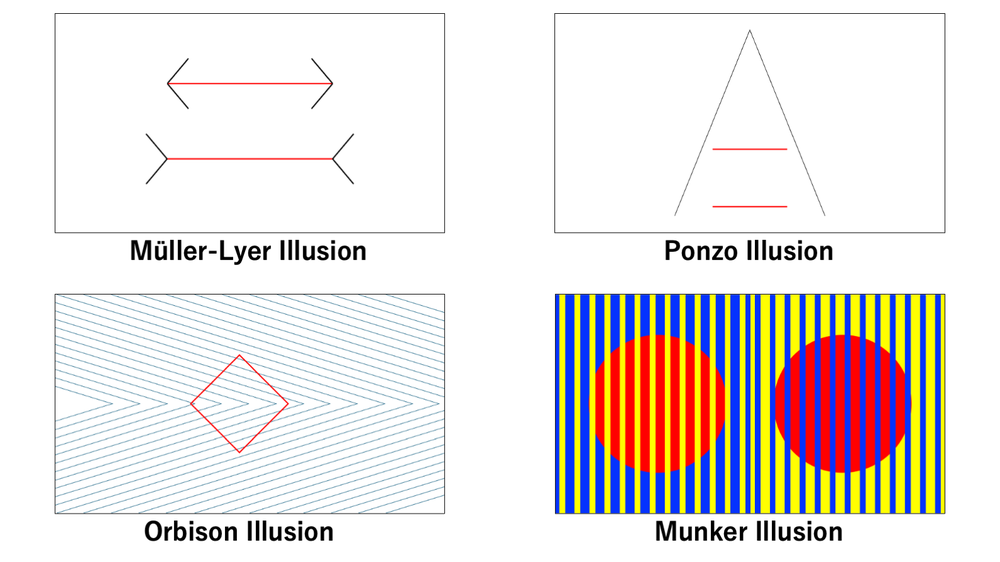}
    \caption{Examples of illusions. The full list is available at \url{https://github.com/ynklab/FILM}.}
    \label{fig:ex_ills}
\end{figure}

%% file: table/result-g.tex
\begin{table*}[th!]
    \centering
    \caption{
    Accuracy for the Genuine Illusion VQA task. There are no human errors regarding apparent features because we excluded images that did not have appropriate apparent features (see Human Evaluation section).
    }
    \label{tab:result-g}
    \vskip 0.12in
    \resizebox{\textwidth}{!}{
    \begin{tabular}{lrrrrrrrrrrrrr} \hline
    \multicolumn{1}{l}{Model}
    & \multicolumn{3}{c}{GPT-4o} 
    & \multicolumn{3}{c}{Claude3.5} 
    & \multicolumn{3}{c}{LLaVA-72b} 
    & \multicolumn{3}{c}{LLaVA-110b} 
    & \multicolumn{1}{c}{Humans} \\ \hline
    Setting        
    & 0-shot & 1-shot & Metacog 
    & 0-shot & 1-shot & Metacog
    & 0-shot & 1-shot & Metacog
    & 0-shot & 1-shot & Metacog
    & --- \\ \hline
    Both Correct    
    & 75.0 & 71.4 & 50.0 
    & 60.7 & 64.3 & 71.4 
    &  0.0 & 7.1  &  0.0
    &  7.1 & 10.7 &  7.1
    & 62.5 \\ \hline
    Only Apparent   
    &  7.1 &  0.0 & 17.9 
    &  7.1 &  0.0 &  7.1 
    & 28.6 & 25.0 & 28.6
    & 25.0 & 14.3 & 25.0
    & 37.5 \\ \hline
    Only Actual     
    & 14.3 & 28.6 & 28.6 
    & 32.1 & 35.7 & 21.4 
    & 35.7 & 35.7 & 35.7
    & 42.9 & 53.6 & 42.9
    &  --- \\ \hline
    Neither Correct  
    &  3.6 &  0.0 &  3.6 
    &  0.0 &  0.0 &  0.0 
    & 35.7 & 32.1 & 35.7
    & 25.0 & 21.4 & 25.0
    &  --- \\ \hline
    \end{tabular}
    }
\end{table*}

%% file: table/result-f.tex
\begin{table*}[th!]
    \centering
    \caption{Accuracy for the Fake Illusion VQA task. Only responses to fake illusions that correspond to genuine illusions where the responses of each model or human are Both Correct are considered.}
    \label{tab:result-f}
    \vskip 0.12in
    \begin{tabular}{lrrrrrrrrr} \hline
    \multicolumn{1}{l}{Model}
    & \multicolumn{3}{c}{GPT-4o}
    & \multicolumn{3}{c}{Claude3.5}
    & \multicolumn{1}{l}{Humans}    \\ \hline
    Setting       & 0-shot& 1-shot&Metacog& 0-shot& 1-shot&Metacog&    ---  \\ \hline
    Both Correct  &  14.3 &  10.0 &  21.4 &   0.0 &  11.1 &   5.0 &  100.0  \\ \hline
    Only Apparent &  81.0 &  85.0 &  78.6 &  94.1 &  83.3 &  95.0 &    0.0  \\ \hline
    Only Actual   &   0.0 &   0.0 &   0.0 &   0.0 &   0.0 &   0.0 &    0.0  \\ \hline
    Neither Correct&   4.8 &   5.0 &   0.0 &   5.9 &   5.6 &   0.0 &    0.0  \\ \hline
    \end{tabular}
\end{table*}

%% file: figure/model-ans-ex.tex
\begin{figure}[t]
    \centering
    \includegraphics[width=8.5cm]{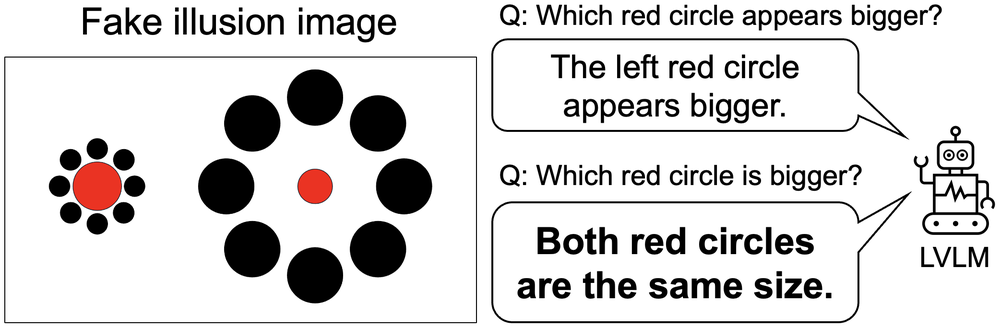}
    \vspace{-1em}
    \caption{
    Example of model responses to the Fake Illusion VQA task. For fake illusion images, the models response correctly to questions about the apparent features but incorrectly to questions about the actual features (bold text).
    }
    \label{fig:model-ans-ex}
\end{figure}

%% file: table/comparison_control-g.tex
\begin{table}[t!]
    \centering
    \caption{
        Comparison of accuracy on Control VQA task corresponding to genuine illusion with those on Genuine Illusion VQA task in zero-shot prompting.
    }
    \label{com_con1}
    \vskip 0.12in
    \resizebox{\columnwidth}{!}{
    \begin{tabular}{lrrrrr}
        \hline
        Model & \multicolumn{2}{c}{GPT-4o}  & \multicolumn{2}{c}{Claude3.5} & Humans \\ 
        \hline
        VQA   & Genuine & Control  & Genuine & Control    & Control\\ 
        \hline
        Both Correct   & 75.0 & 75.0  & 60.7 & 75.0  & 100.0 \\ 
        Only Apparent  &  7.1 &  0.0  &  7.1 &  0.0  &   0.0 \\ 
        Only Actual    & 14.3 & 17.9  & 32.1 & 25.0  &   0.0 \\ 
        Neither Correct &  3.6 &  7.1  &  0.0 &  0.0  &   0.0 \\ 
        \hline
    \end{tabular}
    }
\end{table}

%% file: table/comparison_control-f.tex
\begin{table}[t!]
    \centering
    \caption{
        Comparison of accuracy for the Control VQA task corresponding to fake illusions with accuracy for the Fake Illusion VQA task using zero-shot prompting.
    }
    \label{com_con2}
    \vskip 0.12in
    \resizebox{\columnwidth}{!}{
    \begin{tabular}{lrrrrr}
        \hline
        Model & \multicolumn{2}{c}{GPT-4o}  & \multicolumn{2}{c}{Claude3.5} & Humans \\ 
        \hline
        VQA   & Control & Control  & Control & Control    & Control \\ 
        \hline
        Both Correct   & 14.3 & 52.4  & 0.0  & 11.8  & 100.0 \\ 
        Only Apparent  & 81.0 & 38.1  & 94.1 & 88.2  &   0.0 \\ 
        Only Actual    &  0.0 &  0.0  & 0.0  &  0.0  &   0.0 \\ 
        Neither Correct &  4.8 &  9.5  & 5.9  &  0.0  &   0.0 \\ 
        \hline
    \end{tabular}
    }
\end{table}